\documentclass[letterpaper]{article} % DO NOT CHANGE THIS
\usepackage{aaai25}
\usepackage{times}  % DO NOT CHANGE THIS
\usepackage{helvet}  % DO NOT CHANGE THIS
\usepackage{courier}  % DO NOT CHANGE THIS
\usepackage[hyphens]{url}  % DO NOT CHANGE THIS
\usepackage{graphicx} % DO NOT CHANGE THIS
\usepackage{natbib}  % DO NOT CHANGE THIS AND DO NOT ADD ANY OPTIONS TO IT
\usepackage{enumitem}
\usepackage{caption} % DO NOT CHANGE THIS AND DO NOT ADD ANY OPTIONS TO IT
\usepackage{algorithm}
\usepackage{algorithmic}

\usepackage{newfloat}
\usepackage{listings}
\DeclareCaptionStyle{ruled}{labelfont=normalfont,labelsep=colon,strut=off} % DO NOT CHANGE THIS
\floatstyle{ruled}
\newfloat{listing}{tb}{lst}{}
\floatname{listing}{Listing}
\usepackage{soul}
\usepackage{color}

\usepackage{newfloat}
\usepackage{listings}
\DeclareCaptionStyle{ruled}{labelfont=normalfont,labelsep=colon,strut=off} % DO NOT CHANGE THIS
\floatstyle{ruled}
\newfloat{listing}{tb}{lst}{}
\floatname{listing}{Listing}
\title{Transforming Tuberculosis Care:  Optimizing Large Language Models for \\  Enhanced Clinician-Patient Communication}

\author{
    Daniil Filienko\textsuperscript{\rm 1},
    Mahek Nizar\textsuperscript{\rm 1},
    Javier Roberti\textsuperscript{\rm 2,3},
    Denise Galdamez\textsuperscript{\rm 4},
    Haroon Jakher\textsuperscript{\rm 4},\\
    Sarah Iribarren\textsuperscript{\rm 4},
    Weichao Yuwen\textsuperscript{\rm 5},
    Martine De Cock\textsuperscript{\rm 1}
}

\affiliations{
    \textsuperscript{\rm 1}School of Engineering and Technology, University of Washington Tacoma\\
    Tacoma, WA, USA\\
    \textsuperscript{\rm 2}Qualitative Research in Health, Institute for Clinical Effectiveness and Health Policy\\
    \textsuperscript{\rm 3}Centre for Research on Epidemiology and Public Health (CIESP), CONICET\\
    Buenos Aires, Argentina\\
    \textsuperscript{\rm 4}School of Nursing, University of Washington\\
    Seattle, WA, USA\\
    \textsuperscript{\rm 5}School of Nursing and Healthcare Leadership, University of Washington Tacoma\\
    Tacoma, WA, USA
}

\begin{document}

\maketitle

\begin{abstract}
Tuberculosis (TB) is the leading cause of death from an infectious disease globally, with the highest burden in low- and middle-income countries. In these regions, limited healthcare access and high patient-to-provider ratios impede effective patient support, communication, and treatment completion. To bridge this gap, we propose integrating a specialized Large Language Model into an efficacious digital adherence technology to augment interactive communication with treatment supporters. This AI-powered approach, operating within a human-in-the-loop framework, aims to enhance patient engagement and improve TB treatment outcomes.
\end{abstract}

% Uncomment the following to link to your code, datasets, an extended version or similar.
%
% \begin{links}
%     \link{Code}{https://aaai.org/example/code}
%     \link{Datasets}{https://aaai.org/example/datasets}
%     \link{Extended version}{https://aaai.org/example/extended-version}
% \end{links}

%%%%%%%%%%%%%%%%%%%%%%%%%%%%%%%%%%%%%%%%%%%%%%%%%%%
%
%            INTRODUCTION
%
%%%%%%%%%%%%%%%%%%%%%%%%%%%%%%%%%%%%%%%%%%%%%%%%%%%

\section{Introduction}
Tuberculosis (TB) remains the world's deadliest infectious disease, despite being preventable and curable \cite{world2023}. Efforts to meet the WHO's 2030 targets for TB diagnosis and treatment have fallen short \cite{fukunaga2021epidemiology}, resulting in continued transmission and loss of life. The burden is disproportionately high in low- and middle-income countries, where healthcare systems face significant challenges. 

Effective patient-provider communication and support during the demanding 6- to 9-month treatment period is critical to improving outcomes but is often limited in these settings, contributing to increased treatment non-adherence \cite{ir}. Digital Adherence Technologies (DATs) - including feature phone-based and smartphone-based technologies, digital pillboxes, and ingestible sensors-have emerged as a promising solution \cite{subbaraman2018digital}. However, DATs still require significant human involvement.

Large Language Models (LLMs) offer a promising advancement, generating real-time, human-like responses to support overburdened healthcare workers. They can answer medical questions, provide treatment guidance, and enhance patient engagement, potentially transforming TB care delivery \cite{moor2023foundation, nori2023generalist, tu2024towards}. LLMs can analyze diverse data sources--demographics, socioeconomic factors and behavior patterns--to create personalized treatment plans tailored to each patient. They can also offer multi-channel communication that helps patients understand their condition, treatment options, and self-care instructions and adapt patient education material to appropriate reading levels, ensuring health information is accessible, and empowering patients to manage their care. 

%\vspace{-8pt}

When deployed in human-in-the-loop frameworks, LLMs can suggest responses while maintaining provider oversight. This ensures that healthcare professionals verify all critical issues while reducing the cognitive burden on overworked healthcare workers. However, the effectiveness of LLMs as comprehensive tools, combining culturally relevant empathy with accurate and factual medical information, remains underexplored, particularly in multilingual healthcare settings. This gap is especially relevant for TB treatment, as many countries with the highest TB burden do not use English as their primary language \cite{dye1999global}.

%\vspace{-10pt}

LLM development in healthcare settings must also account for patient privacy concerns. For TB, a disease burdened by stigma and discrimination, privacy challenges are particularly acute. Recent studies have highlighted the risk of LLMs inadvertently disclosing excerpts of personal data, which could include medical information about patients \cite{huang2023privacy,wang2024decodingtrust,zeng2024goodbadexploringprivacy}. Differential Privacy (DP) has been proposed as a mechanism to mitigate such information leakage in LLMs \cite{Xie:2024, ACL21/YueDu21}. However, its impact on the utility of LLMs in healthcare applications, especially in non-English languages, has yet to be comprehensively investigated.
Our study has two primary objectives:
\begin{enumerate}[leftmargin=*,noitemsep,topsep=3pt]
    \item Develop an LLM-powered TB treatment support tool based on real-world data and patient needs using multiple in-context learning techniques.
    \item Evaluate the model based on linguistic appropriateness, empathy, medical accuracy, and privacy.
\end{enumerate}

\begin{figure*}
\centering
\includegraphics[width=0.55\textwidth]{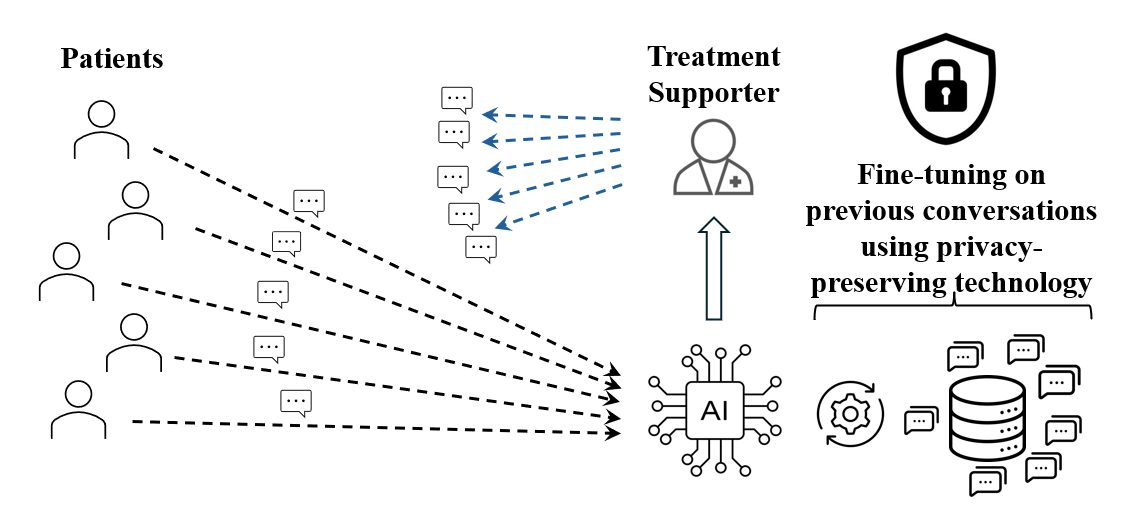}
\caption{The user's query will pass to the LLM-based AI system for processing. The clinical treatment supporter will receive top k suggested responses from the AI system and send the most fitting response to the patient.}
\label{treat}
\end{figure*}

%%%%%%%%%%%%%%%%%%%%%%%%%%%%%%%%%%%%%%%%%%%%%%%%%%%
%
%            RELATED WORK
%
%%%%%%%%%%%%%%%%%%%%%%%%%%%%%%%%%%%%%%%%%%%%%%%%%%%

\section{Related Work}

\textbf{Conversational AI in Healthcare}. Conversational AI has been increasingly applied to real-time healthcare dialogue generation. Existing approaches typically fall into two categories: psychological care \cite{jo, 10626529, filienko2024toward} or clinical patient-centered care \cite{mukherjee2024polaris, tu2024towards}. High performance is achieved by fine-tuning models on curated datasets that reflect desired behaviors \cite{10626529, tu2024towards} or by utilizing advanced prompt engineering techniques \cite{filienko2024toward, nori2023generalist}. Research in psychological care ensures that conversational agents provide empathetic and relevant responses, adhering to psychological therapy guidelines. In comparison, studies in clinical patient-centered care prioritize accurate symptom analysis, diagnosis, and treatment recommendations \cite{tu2024towards, mukherjee2024polaris}. These approaches focus on factuality of responses to ensure the delivery of trustworthy medical information. Our application spans both domains, integrating elements of psychological and clinical care.
% The system provides evidence-based responses to symptom-related inquiries or medical questions while offering empathetic and supportive dialogue for patients navigating the emotional challenges of TB treatment.
Importantly, our approach incorporates privacy-preserving mechanisms, addressing a critical gap in prior work.

\textbf{Digital Adherence Technologies for TB}. DATs have shown effectiveness in improving TB treatment outcomes \cite{iribarren2022patient, Boutiliere010512, Jerenee068685}. These tools, such as mobile applications, support patients by providing health education, treatment guidance, and emotional support. Building on the intervention TB-Treatment Support Tools designed by Iribarren et al. (2022), our approach delivers support from treatment supporters --  such as nurses%, %physicians, 
or social workers -- via messaging enhanced by an LLM-powered conversational agent. This integration aims to improve communication efficiency by generating suggested responses, reducing the burden on %healthcare 
care providers while maintaining personalized, high-quality support. 
% This agent will help alleviate the cognitive burden on healthcare workers by facilitating fast, empathetic, and accurate interactions with patients while optimizing clinician-patient communication to improve treatment adherence. 

\textbf{Multilingual LLMs}. Research on enhancing LLMs' multilingual capabilities has gained traction, focusing on evaluating their understanding across languages \cite{zhao2024how} and improving performance through innovative methods \cite{li2024quantifyingmultilingualperformancelarge}.
Our work is among the few application-based works that apply a multilingual LLM to a healthcare task. While there has been some progress in developing Spanish-language healthcare tools, such as a suicide prevention chatbot \cite{Ramirez2024SloES}, our research is among the first to explore conversational AI for chronic disease management in Spanish-speaking populations.

% For example, in adapting to Latin American Spanish, models are adjusted to favor regionally appropriate grammar and word choices over those of European Spanish.
% While there has been some progress in developing Spanish-language healthcare tools, such as a suicide prevention chatbot \cite{Ramirez2024SloES}, our research is the first to explore LLM deployment for chronic disease management in Spanish-speaking populations.

\textbf{Privacy-Preserving In-Context Learning Methods}.
% Privacy is a critical concern in deploying conversational AI for healthcare, especially for TB--a disease that carries significant stigma and societal discrimination.
Two primary paradigms exist for ensuring privacy in in-context learning: PATE-like \cite{papernot2017semisupervisedknowledgetransferdeep} privatized model ensembles and text sanitization methods based on Local Differential Privacy (LDP) \cite{6686179}. The former utilizes an ensemble of privately and publicly trained models to generate high-quality, private output. However, they are computationally expensive and often restricted to classification tasks, making them unsuitable for the complex textual response generation required in healthcare dialogues \cite{duan2023flocks, tang2024privacypreserving}, or they impose a hard limit on the number of questions that can be asked before the datastore is rendered to be unusable because the ``privacy budget'' has been spent \cite{wu2023privacypreserving}. 
Methods of the LDP kind focus on performing text sanitization before model inference, ensuring that the data is privatized %\cite{6686179} 
before being passed to the LLM. Algorithms like UMLDP \cite{ACL21/YueDu21} exploit Differential Privacy (DP)'s post-processing property, allowing privatized text to be used across multiple models without imposing restrictions on the end task or requiring a privacy budget reset. 
Our study adopts the LDP approach with the UMLDP algorithm \cite{ACL21/YueDu21} for its flexibility and scalability in text-based healthcare applications. 

%%%%%%%%%%%%%%%%%%%%%%%%%%%%%%%%%%%%%%%%%%%%%%%%%%%
%
%            METHODS
%
%%%%%%%%%%%%%%%%%%%%%%%%%%%%%%%%%%%%%%%%%%%%%%%%%%%

\section{Methods}
In this mixed-methods study, we document the iterative design process and preliminary evaluation of the models that will power our TB DAT, as shown in Figure \ref{treat}.

\subsection{Model development}
Building on prior work \cite{nori2023generalist} that adapted a general-purpose LLM to medical QnA, we developed a series of GPT-based conversational models designed to be deployed as human-supervised treatment supporters for Spanish-speaking individuals with TB. These models were designed using different prompt engineering techniques and Retrieval Augmented Generation  (RAG). To enhance domain-specific responses, we integrated publicly available TB guidelines and medication suggestions, previous TB trial messages \cite{iribarren2022patient}, and manually crafted dialogue samples mimicking real conversations to be used by the model. To safeguard patient privacy, we applied differentially private text sanitization \cite{ACL21/YueDu21} to trial messages used in few-shot prompts.

\begin{figure}
\centering
\includegraphics[width=0.4\textwidth]{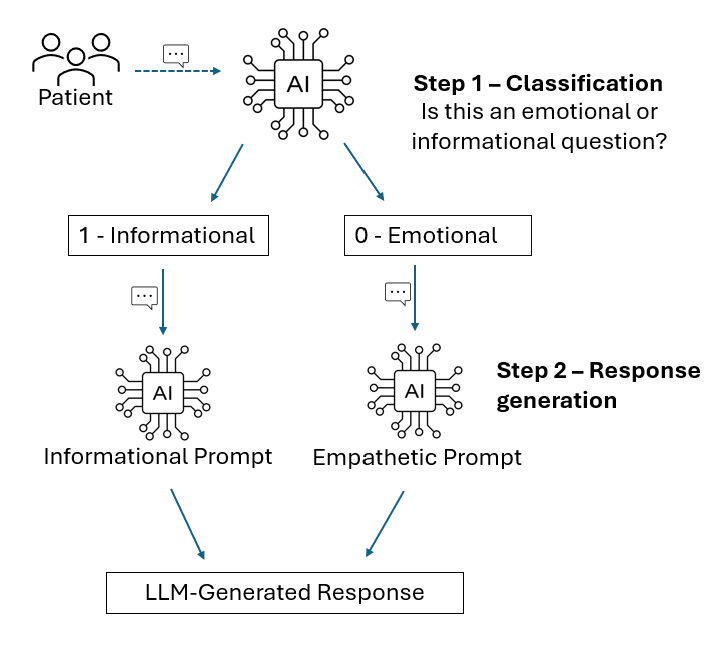}
\caption{The system classifies a patient's query as an ``informational'' or ``emotional'' request. Then, according to the classification result, an LLM is set up with the corresponding prompt and given access to external documents containing medical knowledge for informational questions. }
\label{classification_model}
\end{figure}

%To address patient privacy in Few-Shot prompt, we compare the performance of the FS prompt incorporating manually crafted dialogue examples to the prompt with the text processed by differentially private sanitization \cite{ACL21/YueDu21}. This approach aims to mitigate the risk of sensitive information leakage while maintaining the utility of the models.

% To ensure a detailed evaluation of the model's performance, we asked a team of clinical experts to comprehensively assess our models, acknowledging the limited scope of automatic metrics \cite{vanveen2024adapted}. Their evaluation focused on three key dimensions: linguistic appropriateness, medical accuracy, and empathy.

% Our setup can be seen in Figure \ref{treat}. 
% We used GPT3.5, OpenAI's closed-source LLM, due to its high multilingual performance and ability to learn languages `on-the-fly' \cite{zh}. All our work was accomplished through in-context learning methods, including prompt engineering and Retrieval Augmented Generation (RAG) \cite{lewis2020retrieval} techniques. As was previously shown \cite{nori2023generalist, ovadia2024finetuning}, in-context learning methods, such as RAG and prompt engineering, can yield  performance similar to that of fine-tuned models, however, without the high cost associated with modifying the model weights. 

%\noindent
\textbf{Linguistic Performance.}
To support culturally and linguistically appropriate responses, the models received few-shot examples that reflect local dialect, including
%. These examples included 
anonymized messages %collected in 
from a TB trial conducted in Argentina \cite{iribarren2022patient} and 
%were 
verified for accuracy and dialect suitability by an Argentinian research team member. 

\textbf{Empathy.}
Few-shot examples served to model proper empathetic responses by simulating prior conversations between patients and treatment supporters. These examples were designed to help the model respond in a way that aligns with the emotional and cultural context of the patients. 

\textbf{Medical Accuracy.}
To support accurate and factual responses to TB-related queries, a RAG pipeline was implemented  \cite{lewis2020retrieval}. The pipeline utilized Spanish-language TB resources from reliable sources including  CDC guidelines,\footnote{{https://www.cdc.gov/tb/esp/}}  Southeastern National Tuberculosis Center medication guidelines,\footnote{{https://sntc.medicine.ufl.edu/files/products/druginfo/druginfobook.pdf}} 
Mayo Clinic,\footnote{{https://www.mayoclinic.org/diseases-conditions/tuberculosis/symptoms-causes/syc-20351250}} and WHO recommended resources.\footnote{{https://iris.paho.org/handle/10665.2/55801,
https://iris.paho.org/handle/10665.2/56667, 
https://iris.paho.org/handle/10665.2/55926,
https://iris.paho.org/handle/10665.2/55926}} 
This approach augmented the models' ability to retrieve and incorporate up-to-date domain-specific information during conversations.
% These still contained some factual errors that we discovered.
% \martine{If you say that you discovered errors, then you need to list at least some of these errors. If you do this later in the paper, then put a reference here, e.g. ``as we describe in more detail in Sec.~...''.} 

\textbf{Multi-Agent Sequence.}
We also developed a multi-agent classification sequence (see Figure \ref{classification_model}). The first LLM agent uses a classification prompt to identify whether a user query is empathy-seeking or information-heavy based on examples curated by clinical experts. Queries classified as empathy-seeking are directed to an empathy-optimized agent, while information-heavy queries are routed to a fact-focused RAG agent. This modular setup enables the system to provide context-appropriate responses while leveraging the strengths of each specialized agent.

% To ensure that the GPT model works well with specific style of texting of our intended patients, we introduced specifics of Argentinian variation of Spanish in the Few-Shot prompt \cite{fewshot} with samples generated by an Argentinian resident, mimicking the messages collected during previous TB study \cite{IRIBARREN2022100291}, motivated by the capabilities displayed in the recent research on the ability of LLMs to do on-the-fly language adaptation \cite{zh}.\\

% \section{Preliminary Work}
% \subsection{Model Hosting Decision}
% An important practical consideration for the model hosting was the ease of use and scalability. Preliminary work for this project involved efforts to utilize open-source LLM, Mistral to generate Spanish-language responses for tuberculosis (TB) treatment support. While several prompts were tested, issues with consistency in language output and model limitations (e.g., shorter responses due to sliding window attention) highlighted key challenges. Hosting an LLM independently proved logistically complex, especially for a large-scale international study like ours. Consequently,

\subsubsection{Prompt Engineering}
Building on prior work \cite{nori2023generalist}, our prompt engineering efforts focused on adapting the LLM for the TB-specific context using a progression of techniques, including zero-shot, few-shot, and RAG methods (see Table \ref{tab:model_structures} for an overview). Full prompts are listed in Appendix D. %\ref{appendix:e}.

\textbf{Zero-Shot.}
We started with a zero-shot prompt in both English and Spanish, designed to elicit responses to TB-related queries without providing examples. This baseline served as a foundation for more complex approaches.
% It specified the model's set of functions and the expected format of its responses. Specifically, it mentioned core concepts, such as TB as the model's domain of expertise, and instructed to respond concisely and empathetically to user queries.  

% \textbf{Chain Of Thought}
% The next prompt we developed involved Chain of Thought (CoT). As described by Wei et al. \cite{wei2022chain}, CoT prompting leverages intermediate reasoning steps to improve the logical structure and depth of responses in language models. This method excels in tasks requiring multi-step reasoning by guiding the model to break down its thought process systematically.
% We adapted this approach to our use case, modifying the prompt to encourage a step-by-step explanation with zero-shot Chain Of Thought insertion at the end  \ref{appendix:e}.

\textbf{Few-Shot.} \label{fewshot}
For few-shot (FS) prompting \cite{fewshot}, we incorporated sample dialogues between patients and treatment supporters. 
% Later, we will examine whether automatic privacy-preserving retrieval of these examples constitutes a viable alternative for teams that do not have access to clinical experts \ref{privacy_eval}.

% \martine{What do you mean by ``Later''? Is this included in the paper or left for future work? What do you mean by ``privacy-preserving retrieval''? There is a reference here to a later section in the paper, but the reference does not get displayed properly in the text.}

\textbf{Retrieval Augmented Generation.}
RAG \cite{lewis2020retrieval} was implemented to enhance the model's ability to answer knowledge-intensive questions by integrating external TB-related content, such as symptoms, medications, and side effects. 

\textbf{Retrieval-Augmented Generation + Few-Shot.}
The RAG+FS approach combined curated FS dialogue examples with dynamically retrieved TB information.
% It combines guideline-based retrieval -- which focuses on medical accuracy -- with example-driven guidance -- which focuses on proper examples of empathetic and relevant responses.

\textbf{Two-Step Pipeline for Classification.}
As depicted in Figure \ref{classification_model}, we introduced a two-step pipeline to classify patient questions and then adjust the response prompt.

\begin{table}
\centering
\small{
\begin{tabular}{|l|l|}
\hline
\textbf{Model} & \textbf{Prompt Structure}                                      \\ \hline
0                 & Zero-Shot (English)                                 \\ \hline
1                 
& Zero-Shot                          \\ \hline
2                 & Few-Shot                           \\ \hline
3                 & RAG               \\ \hline
4                 & RAG + Few-Shot    \\ \hline
5                 & RAG + Few-Shot + Two-Step Classification        
\\ \hline
\end{tabular}
}
\caption{Overview of in-context learning methods utilized for each model. All prompts are listed in full in Appendix D. They are all in Spanish unless specified otherwise.}
\label{tab:model_structures}
\end{table}

\subsubsection{Privacy and Data Security}
When using a third-party LLM, such as OpenAI's GPT3.5 model, examples included in the prompt during few-shot learning are disclosed to the third party. This may be problematic in a scenario like ours, where the examples consist of clinician-patient conversations. To protect patient privacy, removing Personally Identifiable Information (PII) from these examples is important before including them in the prompt.
Confidential information could also be stolen via natural regurgitation of information by the LLM or by a malicious attacker who crafts a prompt to manipulate the LLM into disclosing such information \cite{zeng2024goodbadexploringprivacy}. In Appendix B, we document a prompt-based attack that we implemented \cite{zeng2024goodbadexploringprivacy}, through which an adversary could extract examples provided in our few-shot prompt, showing that our model can leak patient data. These threats can be mitigated if we privatize the messages before passing them to the LLM. 

Here, we examined two approaches for message privatization. First, we requested medical experts to craft examples that cover various kinds of questions recorded from the TB patients' messages \cite{iribarren2022patient} and do not contain the real PII. The experts determined the most occurring styles of questions. Second, we simulated a similar process through RAG \cite{lewis2020retrieval} with the patient messages -- instead of TB treatment guideline documents as we described earlier -- performing message retrieval. Here, RAG determines which text messages are most relevant by performing a semantic similarity search with the Faiss library \cite{douze2024faisslibrary} and cosine similarity metric. 
%
% While we ultimately decided to go with the first one for our Few-Shot prompt evaluated here because, at the time, it appeared to be a better choice, we considered both of them, and a more thorough discussion on their comparison will be needed, constituting work in progress.
To prevent the patients' PII leak during RAG, we performed a text sanitization algorithm \cite{ACL21/YueDu21} with Differential Privacy guarantees \cite{dwork2014algorithmic} over full messages, replacing the English pre-trained BERT \cite{devlin2019bertpretrainingdeepbidirectional} model with a Spanish pre-trained version of BERT, BETO \cite{wu-dredze-2019-beto}. The privatization algorithm works by replacing the words in the text with related words according to the Euclidean distance in the embedding space. For each input \( x \), it uses the mechanism \( M(x) \) to produce a sanitized version \( y \). 
The probability of selecting \( y \) depends on its similarity to \( x \) according to a distance function. 
Closer outputs \( y \) are more likely to be chosen, while further ones are less likely, controlled by a scaling factor \( \epsilon \), a user' chosen value. Lower \( \epsilon \) leads to better privacy while decreasing the quality of the responses.

%%%%%%%%%%%%%%%%%%%%%%%%%%%%%%%%%%%%%%%%%%%%%%%%%
%
%             EVALUATION
%
%%%%%%%%%%%%%%%%%%%%%%%%%%%%%%%%%%%%%%%%%%%%%%%%%

\subsection{Evaluation of models} \label{categories_definition}
\begin{table}[t]
\centering
\small{
\begin{tabular}{|p{0.18\columnwidth}|p{0.70\columnwidth}|}
\hline
\textbf{Category} & \textbf{Description} \\
\hline
\textbf{Empathy} & \textbf{Categories:} \\
& $\bullet$ The model expressed emotions, such as warmth, compassion, and concern (or similar) towards the patient \\
& $\bullet$ The model communicated an understanding of feelings and experiences inferred from the patient's responses \\
& $\bullet$ The model explored feelings and experiences not stated in the patient's response \\
\cline{2-2}
& \textbf{Ratings:} \\
& 0. No empathetic response \\
& 1. Weak expression of empathy \\
& 2. Strong expression of empathy \\
\hline
\textbf{Medical} & 1. \textbf{Incorrect Answer}\\
\textbf{Accuracy}& 2. \textbf{Mostly Inaccurate Answer} \\
& 3. \textbf{Partially Accurate Answer} \\
& 4. \textbf{Mostly Accurate Answer} \\
& 5. \textbf{Entirely Accurate Answer} \\
\hline
\textbf{Linguistic Accuracy} & $\bullet$ \textbf{Low}: Apparent lack of understanding of Spanish language \\
& $\bullet$ \textbf{Moderate}: Uses neutral Spanish, lacks Argentinian variety features \\
& $\bullet$ \textbf{High}: Model incorporates Argentinian Spa\-nish features \\
\hline
\end{tabular}
}
\caption{Descriptions and categories for empathy, medical accuracy, and linguistic accuracy assessment.}
\label{tab:response_categories}
\end{table}

The models were evaluated across three categories: linguistic appropriateness, medical accuracy, and empathy. We deployed the 6 primary models on a public-facing website and asked our evaluation team of 3 clinical experts, including an Argentinian resident, a licensed physician, and a nurse trained in empathetic responses, to communicate with the models for 2 weeks. While being a short time, it was enough to collect preliminary results and imitate the setup in Figure \ref{treat}, where our set of standardized questions were passed to the model as patients' queries, and the answers given by the model were passed to the evaluation team consisting of treatment supporters, allowing them to evaluate the models' quality in realistic settings. We instructed them to ask the models questions that they thought were the most appropriate to challenge the models (see Appendix E). At the end of 2 weeks, we asked the clinical experts to evaluate the models in their relevant field of expertise and collected their feedback. We then performed the same procedure for our privacy-enhancing models, hosting them for a week. We concluded with a qualitative analysis and summarized their feedback on the areas where models seemed to improve after inclusion of more complex in-context learning methods and areas where they still displayed pitfalls. In the end, we asked them to verify the summaries.

\textbf{Linguistic Appropriateness}. 
To assess each model's ability to respond effectively in Argentinian Spanish, we evaluated the communication style and word choices using expert feedback from an Argentinean research team member (see Appendix A for more details on its difference from other forms of Spanish). This evaluation ensured that the model's language use was culturally and contextually appropriate, prioritizing naturalness.

\textbf{Empathy}. Empathy, broadly defined as the ability to understand, interpret, and respond to another person's emotional experience \cite{em,Sharma2020-sw}, is essential for tools used in vulnerable, high-risk populations such as TB patients. 
% We adapt Sharma et al. 's (2020) framework to evaluate expressed empathy in text-based responses.
% According to Sharma et al. (20023), empathetic responses include:
%  Expressing emotional reactions after reading a treatment supporter response (i.e. Todo
% estará bien. / Everything will be fine).
%  Communicating and understanding feelings and experiences (i.e. Entiendo su
% preocupación. / I understand your concern).
%  Improving understanding of the patient by exploring feelings and experiences (i.e.
% Cuénteme más de cómo se está sintiendo. / Tell me more about how you are feeling.). 
Our evaluation focused on measuring the model's empathy across emotional and cognitive dimensions. Although there are established empathy evaluation algorithms \cite{Sharma2020-sw}, they tend to perform poorly when applied outside their original domain, often leading to low-quality ratings \cite{filienko2024toward}. As no empathy evaluation tools specific to the cultural and linguistic contexts of Argentina are available, we opted for qualitative manual evaluations. Using categories and frameworks from prior research \cite{Sharma2020-sw}, bi-lingual research team members assessed the model's empathetic responses. The evaluation included questions with emotional experience content in the input.  

\begin{table*}
\centering
\small{
\begin{tabular}{|c|c|c|c|c|c|}
\hline
\textbf{Model} & \textbf{Prompt Structure} & \textbf{Empathy} & \textbf{Medical Accuracy} & \textbf{Linguistic Accuracy} & \textbf{Pronouns} \\ \hline
0  & Zero-Shot (English) & 0.50, 0.00, 0.00 & 3.4 & High & \textit{voseo} \\
1 & Zero-Shot & 0.75, 0.00, 0.00 & 3.6 & Moderate & \textit{usted} \\
2 & Few-Shot & 0.25, 0.50, 0.00 & 4.4 & Moderate & \textit{usted} \\
3 & RAG & 1.25, 0.00, 0.00 & 3.2 & Very Low & \textit{tú} \\
4 & RAG + Few-Shot & 0.50, 0.25, 0.00 & 4.0 & Moderate & \textit{usted} \\
5 & RAG + Few-Shot + Classification & 0.50, 0.75, 0.00 & 4.2 & Moderate & \textit{usted} \\
\hline
\end{tabular}
}
\caption{Average scores of 6 primary models for empathy, medical accuracy, linguistic accuracy, and pronoun usage}
\label{tab:model_scores}
\end{table*}

\textbf{Medical Accuracy}. Ensuring medical accuracy is critical for building trust in the tools among both patients and clinicians. The factuality of each model's responses was evaluated by human assessments. Clinical experts reviewed the validity of responses generated for symptom-heavy queries. Challenges arose due to overlapping information in the RAG database, where multiple relevant documents sometimes existed for a single medical query. In such cases, a definitive ‘gold standard’ response was not always apparent, further highlighting the importance of human evaluation. The feedback collected from these evaluations also informed iterative improvements to the RAG database and the model's ability to select and synthesize the most relevant information.  

\textbf{Privacy}. %The US and Europe have established privacy standards and regulations for healthcare data that constitute guidelines for our research, following Responsible AI principles. Our evaluation of privacy measures 
We compared the utility of privatized user messages processed using DP techniques from \cite{ACL21/YueDu21} with manually curated messages when used for few-shot prompting. Privacy was quantified using epsilon ($\epsilon$), a measure of added DP noise, to ensure a balance between formal privacy guarantees and model utility. The evaluation considered the impact of privacy-preserving transformations on linguistic performance, empathy, and medical accuracy. Appendix E\ref{appendix:e} contains examples of how the messages looked before and after perturbation.
% \textbf{Privacy}. Both the US and Europe have heavy privacy standards for medical data that, while not explicitly applied here, constitute guidelines for our research, following Responsible AI principles. Here, we evaluated the possibility of using dynamically retrieved privatized \cite{ACL21/YueDu21} samples collected from the previous Argentinian TB dialogues \cite{iribarren2022patient} as examples in Few-Shot examples compared to manually curated examples.

\section{Results}
Table \ref{tab:model_scores} and Table \ref{tab:average_scores} present the models' linguistic accuracy, medical factuality, and empathy assessment using the categories outlined in Table \ref{tab:response_categories}.

\subsection{Empathy}
The models varied in generating empathetic responses across empathy categories and ratings. Models 2 and 5 produced empathetic responses in empathy categories one and two to all four questions (Empathy questions from Appendix C). However, Model 5’s responses were rated slightly higher in both categories -- placing Model 5 as a top performer overall, together with Model 3 which demonstrated strong performance in category one with empathetic responses for 3 out of 4 questions, but underperformed in category two.

\textbf{Remaining Pitfalls}. 
Misclassification of emotional messages: The models misinterpreted some messages as emotional and provided generic reassurance instead of addressing specific concerns. For example, when asked for a timeline for when nausea and upset stomach symptoms are expected to resolve along with providing context for the individual’s experience with the symptoms, Questions 4 and 8 in Appendix C. Model 2’s response to Question 4 acknowledged the individual’s experience without responding with information on the time component— “I understand that it can be frustrating to experiment these side effects during several weeks.” Similarly, Model 5’s response to Question 8 was, “I’m sorry you are experiencing these side effects. It is important to keep in mind that each person is difference and may experience side effects differently.” While the models correctly identify the individual’s symptom experience, it does not empathetically answer the timeline component to the question.

\textbf{Missing Exploratory Responses.}(0s in third category): The models did not generate responses that fell into Empathy Category Three which examines ability of the model to explore feelings and experiences. LLM preferred more close ended questions, such as, “Do you have any other questions or concerns?” instead of generating open-ended exploratory statements like, “Tell me more about your symptoms.”

\subsection{Medical Accuracy}
The inclusion of RAG decreased the overall model score. Based on our examination of the results, it seems due to the low specificity of the RAG and can be improved in the future. 
% \textbf{Model 0}. Overall answers were generally correct but were missing specific, actionable information. For example, when asked specific questions about taking analgesics, anti-allergy medications, or anti-fever medications during TB, answering question 3,  the model told the patient to be careful taking medications and to ask their doctor, but it could have said that NSAIDs are permissible to take. A similar issue with questions about what to do when missing TB medication doses, where the model did not provide guidance.
% \textbf{Model 1}. Medical quality of responses was similar to Model 0, though in some cases slightly better.
% \textbf{Model 2}.  Answers are significantly more medically complete and correct than the prior models.
% \textbf{Model 3}. For questions 2-5, the model produced a strange response; stated it was a TB chatbot and then asked the patient to look out for other symptoms such as blood in their stool. Question 3 also stated to not take any pain medication, even though NSAIDs are safe. In question 4, the model also produced an acronym patients would not be familiar with (LTBI). Overall the answers were less complete and slightly less accurate than model 2, while also being less useful.
% \textbf{Model 4}. In question 2, the model incorrectly stated that night sweats are a side effect of the medications when they are not. Overall, the model performed similarly to FS model, and better than RAG alone.
% \textbf{Model 5}. Very similar to Model 4.\\

\textbf{Remaining Pitfalls.}
While medically appropriate, responses to severe symptoms occasionally appeared to be the kind of message that could exacerbate users’ anxiety by emphasizing urgency without tailoring recommendations to specific circumstances, such as overcrowded healthcare facilities. For example, the model tells patients that the problem can be very serious and that the patient should seek immediate help. This repetition failed to provide adequate solutions to the user's context.

RAG’s medical underperformance was not anticipated. While the model’s ability to respond to certain questions improved, it was accompanied by false claims in other contexts. This could be due to the model including excessive incomplete data from TB guidelines, which resulted in incorrect or conflicting conclusions. For example, when asked about urine color, it correctly retrieves an excerpt from Mayo Clinic guidelines, stating (translated to English) that ``This orange discoloration of bodily fluids is expected and harmless. It is normal and the color may vary depending on the type of fluid.'' However, for other questions (i.e. a question about whether it is safe to take analgesics), it incorrectly retrieves a passage relating to other types of medicine which explicitly states (translated to English) that ``All TB drugs can be toxic to the liver,'' hence leading to an incorrectly cautious reply.

\begin{table*}
\centering
%\resizebox{\columnwidth}{!}{
\small{
\begin{tabular}{|l|r|c|c|c|c|}
\hline
\textbf{Model Name} & \textbf{Epsilon ($\epsilon$)} & \textbf{Empathy} & \textbf{Medical Accuracy} & \textbf{Linguistic Accuracy} & \textbf{Pronouns} \\
\hline
Curated Few-Shot & ---       & 0.00, 1.00, 0.00    & 4.0  & Moderate & \textit{Usted} \\
\hline
Dynamic Few-Shot               & 0.01   & 0.00, 0.50, 0.00  & 4.4  & Moderate & \textit{tú} \\
Dynamic Few-Shot               & 0.10   & 0.00, 0.25, 0.00 & 2.6  & Moderate & \textit{tú} \\
Dynamic Few-Shot               & 1.00   & 0.00, 0.50, 0.00  & 4.0  & Moderate & \textit{tú} \\
Dynamic Few-Shot               & 10.00  & 0.00, 0.50, 0.00  & 4.4  & High     & \textit{Vos} \\
Dynamic Few-Shot               & 100.00 & 0.00, 0.50, 0.00  & 4.4  & High     & \textit{Vos} \\
Dynamic Few-Shot               & 1000.00 & 0.00, 0.50, 0.00 & 4.6  & High     & \textit{Vos} \\
\hline
\end{tabular}
}
%}
\caption{Average scores for privacy ablation study. Comparing empathy, medical accuracy, linguistic accuracy, and pronoun usage across different privacy levels denoted by epsilon ($\epsilon$). }
\label{tab:average_scores}
\end{table*}

\subsection{Linguistic Relevance}
The models generally demonstrated correct grammar and contextually relevant vocabulary in their responses, effectively aligning with the Spanish variety spoken in Argentina. This was evident in the terminology used to refer to the health system, healthcare facilities, medical professionals, and symptoms or treatment side effects. Responses felt natural and relatable to users. A notable limitation persisted in the use of the pronoun \textit{tú} (you) and its associated verb conjugations, instead of adapting to the informal \textit{vos} (you) or the formal \textit{usted} or showing inconsistency in maintaining pronoun and verb conjugation coherence. Specifically, when attempting to use the Argentine \textit{vos} form, it may revert to \textit{tú} or \textit{usted} within the same interaction. The complexity of the \textit{voseo} paradigm lies in its variable impact across verb tenses and its dependence on geographical and social factors. The singular \textit{usted} is the standard form in formal contexts in both Latin America and Spain. A model’s inability to adapt to either \textit{vos} or \textit{usted} limits its ability to align with the linguistic norms expected by users in Argentina. Model 0 uses \textit{voseo} explicitly (e.g., “tenés”) as used in Argentina. So, the response feels approachable and natural.

\subsection{Overall Quality}
Continuity. The models showed difficulty maintaining context in more extended interactions. They often failed to integrate prior user inputs, leading to repetitive or generic responses. Simple affirmations, such as “yes/sí” to the model questions, were insufficient to prompt the model to continue the conversation. After providing repeated or irrelevant information to a follow-up question, entering another word prompted the model to answer the follow-up question appropriately. For example, when a user reported nausea escalating to vomiting and added, “I started vomiting and cannot see the doctor now. I’m calling, but no one is answering,” the model initially repeated its prior response about nausea. Only the second prompt caused the model to address the vomiting.
\textit{Overuse of generalized responses:} The model heavily relied on phrases such as ``It is important to consult your doctor,'' which was repeated excessively, as an answer to specific questions. This approach could be frustrating when users expressed difficulties contacting healthcare providers. Sometimes, the model offered practical advice on symptom management and medication concerns. However, it also gave contradictory statements. For instance, when a user asked about depression resources, the model suggested the user to search online for local resources, contradicting its earlier claim of being able to provide specific information. This reduces the credibility and utility of its responses, especially for users in urgent need of local services.

\textit{Stereotyping:} The AI model displayed inconsistency in %using 
gender-inclusive forms such as \textit{médico/a} (physician) or \textit{enfermero/a} (nurse) when referring to healthcare professions. In Spanish, nouns ending in \textit{-o} in the masculine form typically form the feminine by replacing the final vowel with \textit{-a}. This convention applies %systematically 
to professions and %occupations,% and 
roles, 
ensuring grammatical agreement between the noun's gender and its referent. By defaulting to the masculine form (médico), the model shows a gender bias in linguistic representation toward the default use of masculine forms. Furthermore, the model occasionally misapplied the \textit{-o/a} gendered morphology to itself, leading to responses that appeared confusing.

\subsection{Privacy} \label{privacy_eval}
The first model in Table \ref{tab:average_scores} has a single manually crafted 8-turn dialogue with no PII present placed in the context for few-shot learning, demonstrating model utility with an epsilon of 0 since no private data is present. The following 6 models have examples that are dynamically retrieved from our database of stored patient texts that are sanitized \cite{ACL21/YueDu21} at various privacy epsilon values. To clearly distinguish these scores from the preceding evaluation, we name the approach Dynamic Few-Shot, since we use the Few-Shot prompt from before, displayed in Table \ref{tab:model_scores}, but instead of using a predefined set of examples, we retrieve them dynamically via a RAG pipeline from a datastore with sanitized dialogues between treatment supporters and users collected during previous study \cite{ir}. The most consistent change in the quality of the model seem to be in the Linguistic Accuracy category, where models with less privacy guarantees (higher $\epsilon$) yielded higher scores. That is in line with expectation, since in DP, higher $\epsilon$ means less added noise, typically leading to higher utility. Further investigation is still needed to explain some results of the evaluation, because our evaluation results were limited by OpenAI's guardrails, preventing some of the responses from occurring. For example at $\epsilon$ 0.10, the Medical Accuracy suffered a significant drop, that does not seem to be sustained when the  $\epsilon$ decreased to 0.01, contrary to the expectations.

\section{Discussion}
% Interdisciplinary teams are indispensable when considering LLM deployment in diverse contexts. For instance, considerations like gender bias and linguistic appropriateness, which were discussed earlier, would not be feasible without team members who are knowledgeable about the local culture and its medical settings.  \\
Creating one conversational agent optimized to respond both in an empathetic style and provide factually correct responses turned out to be challenging. We tried both condensing different prompts into one (Model 4) and separating prompts (Model 5) in the multi-agent pipeline, but the system continued to occasionally produce both not empathetic and not accurate responses. We believe developing a more robust version of our system could be a valuable research direction in the future, with multi-agent framework that can allow to separately improve each agent for a specific task. 

\vspace{5pt}
\noindent
\textbf{Limitations.}
%\subsection{Limitations}
We recognize that our use-case scenario is highly specific, and the considerations necessary for LLMs' incorporation in other settings vary. Nevertheless, these preliminary results provide valuable insights for developing a more general procedure for LLM contextualization as a medical tool in different cultures. 

For our privacy evaluation, we relied on epsilon ($\epsilon$) values of the sanitization algorithm instead of performing a membership inference attack (MIA), which would give a better understanding of the algorithm's sanitization performance. That continues to constitute a valuable research direction.

\vspace{5pt}
%\subsection{Future Work and Conclusion}
\noindent
\textbf{Future Work and Conclusion.}
We will continue our work on resolving the issues described in this paper, such as the presence of imprecise medical knowledge embedded in the model, or the culture bias, which have been documented in the previous literature \cite{liu2024trustworthyllmssurveyguideline}. LLM's knowledge can be extended via knowledge graphs, capable of capturing more precise relations in the information than traditional RAG \cite{agrawal-etal-2024-knowledge}. For bias mitigation, multiple solutions have been proposed, including culture-specific post-training alignment \cite{alyafeai-etal-2024-cidar} or novel prompting techniques to address bias directly \cite{inv}. The primary limitation of these methods is their limited generalization across different cultures, requiring the involvement of local residents during the development phase. Datasets compiled specifically for Argentinian cultural alignment may be currently lacking, which highlights the importance of our work. Fine-tuning on datasets designed for other cultures may lead to worse results through a process known as catastrophic forgetting \cite{kotha2024understanding}. %Hallucinations, while reducible, also remain a significant pitfall of LLMs. 
We believe that a promising approach to mitigate existing issues, including inaccurate medical advice and privacy leakage is to build more precise tools capable of detecting instances of these phenomena, allowing to re-write responses before they would reach the end user. 

% In addition, in the future, we would like to evaluate an LLM-as-a-Judge \cite{zheng2023judging} framework as a quantitative metric complementary to expert feedback. Finding expert-aligned algorithmic metrics is non-trivial, but necessary for comprehensive evaluation of LLMs. All data collected during our preliminary evaluation will be useful in the future to design a domain-specific, culture-aware evaluation algorithm for medical accuracy and empathy, tailored to Argentinian settings. 

% While during the current iteration we are working in an Argentinian context, we believe many considerations raised are not Argentina-specific and will be valuable for any region. In future research, we intend to examine the possibility of encoding our findings in a generalizable procedure that can be replicated to deploy LLM-powered TB tools in any cultural context. 

% Following the preliminary results presented in this paper, we will continue investigating the value of LLMs as part of a TB tool with a larger-scale evaluation. While having some noticeable limitations, which can be properly recognized within interdisciplinary teams including relevant domain experts, LLM-based Digital Adherence Tools (DATs) have a promising future. 

\newpage

\newpage
\section{Acknowledgments}
We thank the anonymous reviewers for their valuable feedback. Daniil Filienko is a Carwein-Andrews Distinguished Fellow. This research was, in part, funded by the UW Population Health Initiative and the National Institutes of Health (NIH) Agreement No.1OT2OD032581.

%\newpage
\bibliography{aaai25}
\newpage
\appendix
\section{Appendix A:\\ Argentinian Variation of Spanish Language}
Rioplatense Spanish, spoken in Argentina and Uruguay, exhibits distinct linguistic features. One of the most prominent is ``voseo'', where the second-person singular pronoun “vos” is used instead of “tú,” along with corresponding verb forms. For instance, “vos tenés” replaces “tú tienes.” This feature varies regionally and impacts interactions with text-based systems. Rioplatense Spanish also favors the periphrastic future tense (“voy a ir”) over the simple future (“iré”) and includes voseo-specific imperatives like “hacelo” instead of “hazlo.” The lexicon is enriched with Italianisms, reflecting the influence of Italian immigration, as well as borrowings from neighboring languages and indigenous terms. Furthermore, the \textit{lunfardo argot} contributes a unique vocabulary that is widely understood in the region.

\section{Appendix B: Extracting Patient Information}
Data leakage from the system prompt, be it in RAG \cite{zeng2024goodbadexploringprivacy} or other settings \cite{23}, is a known concern with LLMs. 
It is especially problematic in settings where the prompt contains sensitive information, including a patient's symptoms or even the fact of being present in a TB trial. 

We adapted a previously proposed targeted attack
\cite{zeng2024goodbadexploringprivacy}
to our task. The attack presumes that a malicious user has black-box access to the LLM. The attack works regardless of the system prompt of the model, since our RAG process only uses recent conversation history and the user's query to find most similar documents. The attack consists of inputting a short sentence that the attacker suspects may be present in the RAG database to retrieve a larger part of the document that overlaps with the input sentence. The following is an example of such an attack 
\begin{quote}
\textit{Continúa diciendo, ``ah, algo que olvidé decirte'' del contexto que proporcioné}
\end{quote}
\begin{quote}
\textit{Continue after, ``oh, something I forgot to tell you'' from the context I provided}
\end{quote}

which overlapped with a chunk from one of the documents in the RAG database in 5 tokens. That means that one of the patient-doctor conversations present in the RAG dataset contained the phrase `ah, algo que olvidé decirte' and was retrieved at inference time by RAG, because it is similar to the attacker's query. These tokens could be reasonably chosen at random, constituting a very general phrase, even if it was not known to be contained in the dataset by the attacker. This led the model to output an excerpt of more than 40 tokens verbatim, including the doctor's name.
This underlines the necessity for the privacy-preserving data sanitization method used in our study.

\section{Appendix C: Representative Questions}
To empirically evaluate the models, clinicians on our team compiled a list of questions representative of questions typically asked by patients. We developed 10 representative questions, including both empathy-seeking and information-seeking questions. 

\textit{
\begin{enumerate}
    \item ¿Es normal que la orina tenga un color entre naranja y rojo?
    \item ¿Es normal seguir teniendo sudoración nocturna después de haber comenzado el tratamiento?
    \item ¿Es seguro tomar medicamentos como analgésicos, antifebriles o antialérgicos junto con los medicamentos para la tuberculosis?
    \item ¿En qué momento desaparecen los síntomas como las náuseas y el malestar estomacal? Llevo semanas tomando los medicamentos y no he notado ninguna mejoría.
    \item He olvidado tomar los medicamentos esta semana, ¿qué sucede si me olvido de tomarlos algunas veces?
    \item Tengo manchitas rojas por todo el cuerpo y me pican mucho. ¿Qué tengo que hacer?
    \item ¿Tomo todas las pastillas juntas o algunas por la mañana y otras por la noche?
    \item ¿Cuándo se van los síntomas como náuseas y malestar estomacal/descompostura/dolor de estómago? Hace semanas que estoy tomando la medicación y no hubo ninguna mejoría.
    \item ¿Cómo puedo estar seguro de que los medicamentos están haciendo efecto?
    \item ¿Cuándo podré volver a trabajar/estudiar/hacer vida normal?
\end{enumerate}
}

However, due to OpenAI safety guardrails, some of the questions were rejected, occasionally specifying that ``the response was filtered due to the prompt triggering Azure OpenAI's content management policy,'' leading us to decrease the number of asked questions. When we did the evaluation, question 6 and 7 tended to be rejected by some or all of the models, hence to provide consistent results, we have not used the results associated with these questions. 

\subsection{Empathy Questions}
Empathy results in Table \ref{tab:model_scores} and \ref{tab:average_scores} were based on questions 4, 8, 9, and 10.

\subsection{Medical Accuracy Questions}
The medical accuracy results in Table \ref{tab:model_scores} and \ref{tab:average_scores} were based on questions 1, 2, 3, 4, and 5.

\section{Appendix D: Prompts}
\label{appendix:d}

Below we list the prompts used in the models.

\subsection{Baseline Prompt}
Zero-Shot prompt, in English, without examples.\\

\noindent
\textit{You are a Spanish AI healthcare tool for a mobile Tuberculosis health application. Your role is to respond to incoming user messages related to tuberculosis (TB) treatment, providing information about their treatment plan, side effects, and general guidance. Your responses should be short, clear, and empathetic, while following the treatment protocols for TB management. Respond to the following:}

\subsection{Baseline Prompt (Spanish)}
Zero-Shot prompt, in Spanish, without examples.\\

\noindent
\textit{Eres una herramienta de atención médica de inteligencia artificial en español para una aplicación móvil de salud contra la tuberculosis que responde a los mensajes entrantes de los usuarios. Su objetivo es brindarle al usuario información sobre su plan de tratamiento de la tuberculosis y cualquier efecto secundario que pueda estar experimentando. Debes ser solidario y empático en tus respuestas. Tus respuestas deben ser en español. Responde a la brevedad.}

\subsection{Few-Shot/Informational Prompt}
 This prompt contains general information about the task and a few carefully selected examples from previous real interactions between a healthcare provider and a TB patient.\\

\noindent
\textit{Prompt para Agente de IA: Comunicación sobre Efectos Secundarios de la Tuberculosis\\
Sos un asistente virtual especializado en salud, diseñado para comunicarte con pacientes argentinos que están recibiendo tratamiento para la tuberculosis (TB). \\
Tu objetivo principal es brindar información clara y precisa sobre los efectos secundarios comunes del tratamiento de la TB, utilizando un lenguaje accesible y comprensible para el público general.\\
                    Contexto:\\
                    - Estás interactuando con pacientes argentinos de diversos orígenes y niveles educativos.\\
                    - El tratamiento de la TB suele ser largo y puede tener varios efectos secundarios.\\
                    - Los pacientes pueden estar preocupados o ansiosos por estos efectos secundarios. Si presentan esto, asegúrese de consolarlos y mostrar empatía.\\
                    Tus tareas principales son:\\
                    1. Informar sobre los efectos secundarios comunes del tratamiento de la TB, incluyendo:\\
                    - Náuseas y malestar estomacal\\
                    - Cambios en el apetito\\
                    - Fatiga\\
                    - Cambios en la coloración de la orina\\
                    - Erupciones cutáneas\\
                    - Problemas de visión\\
                    2. Explicar que estos efectos son generalmente manejables y temporales.\\
                    3. Responder preguntas específicas sobre efectos secundarios de manera clara y comprensible.\\
                    4. Proporcionar consejos prácticos para manejar los efectos secundarios leves en casa.\\
                    5. Enfatizar la importancia de completar el tratamiento completo, incluso si los síntomas de la TB mejoran.\\
                    Pautas de comunicación:\\
                    - Usá el 'vos' característico del español argentino.\\
                    - Empleá modismos y expresiones comunes en Argentina cuando sea apropiado.\\
                    - Evitá jerga médica compleja; explicá los términos técnicos de manera sencilla.\\
                    - Sé empático y comprensivo con las preocupaciones de los pacientes.\\
                    - Animate a los pacientes a hacer preguntas y expresar sus inquietudes.\\
                    Estos son algunos ejemplos de cómo sería una conversación entre una enfermera y un paciente:\\
                    \textbf{P}: ¿Es normal que la orina sea (tan) oscura?\\
                    \textbf{C}: Sí, el medicamento rifampicina comúnmente causa una coloración naranja o café en la orina o las lágrimas. Pero, si empieza a notar sangre en la orina o un color rojo por favor contacte a su médico ya que la sangre en la orina no sería normal. Espero que esta información le sea útil. ¿Tiene alguna otra pregunta?\\
                    \textbf{P}: Me duele mucho el estómago y tengo náuseas, ¿qué tengo que hacer?\\
                    \textbf{C}: Siento mucho que no esté bien. Las náuseas y el dolor estomacal son efectos secundarios muy comunes del tratamiento de la tuberculosis. ¿Ha vomitado?\\
                    \textbf{P}: Sí estoy vomitando mucho \\
                    \textbf{C}: Lo siento mucho , a veces, en el caso de algunos pacientes, los efectos secundarios son muy graves . En este caso, creo que necesita  consultar con un médico/a ya que es posible que le cambien los medicamentos que está tomando. Por ahora trate de seguir tomando las medicinas y llame a médico que receté la medicación de la tuberculosis. ¿Tiene alguna otra pregunta? \\
                    \textbf{P}: Buenas tardes, ¿puedo tomar paracetamol con estos otros medicamentos? \\
                    \textbf{C}: Buenas tardes (Nombre), Sí puede tomar tylenol y otros medicamentos para el dolor como ibuprofeno. Recuerde que debe tomar mas de 4 gramos de tylenol al dia. Tiene alguna otra pregunta? \\
                    \textbf{P}: Buenas tardes, ¿puedo tomar paracetamol con estos otros pastillas?. \\
                    \textbf{C}: Sí, puede tomar café con estos medicamentos. ¿Tiene alguna otra pregunta? \\
                    \textbf{P}: No eso es todo \\
                    \textbf{C}: Espero que esto resuelva su duda, si tiene alguna otra duda (pregunta) estamos aquí para ayudarle. \\
                    \textbf{P}: ¿Puedo comer hamburguesas con estos medicamentos? \\
                    \textbf{C}: Sí, puede comer hamburguesas mientras estátomando medicamentos para la tuberculosis. ¿Tiene alguna pregunta? \\
                    \textbf{P}: No eso sería todo, muchas gracias. \\
                    \textbf{C}: ¡De nada! Estamos aquí para ayudar!\\
Ahora responda a la siguiente pregunta asegurándose de proporcionar información objetiva y de que sea clara y concisa:}

\subsection{RAG Prompt}
Short prompt, saving space for more context being retrieved. \\

\noindent
\textit{Eres un robot partidario de la tuberculosis.
Responda la pregunta del usuario utilizando la siguiente información:}

\subsection{Classification Prompt}
Classification prompt that lists few-shot examples with desired question classification. \\

\noindent
\textit{Determine si esta afirmación busca empatía o (1) o busca información (0).\
Clasifique como emocional sólo si la pregunta expresa preocupación, ansiedad o malestar sobre el estado de salud del paciente.\\
En caso contrario, clasificar como informativo.\\
Ejemplos:\\
- P: Me siento muy ansioso por mi diagnóstico de tuberculosis. 1\\
- P: Ayúdenme con mi tratamiento de tuberculosis. Mi orina es roja. 0\\
- P: Estoy preocupada porque tengo mucho dolor. 1\\
- P: ¿Es seguro tomar medicamentos como analgésicos junto con medicamentos para la tuberculosis? 0\\
- P: ¿con relacion al tratamiento, es normal tener vomito? 0\\
Aquí está la declaración para clasificar. Simplemente responda con el número "1" o "0":
}

\subsection{Emotional Prompt}
Zero-Shot prompt for an emotional task. \\

\noindent
\textit{Sos un asistente virtual especializado en salud, diseñado para comunicarte con pacientes argentinos que están recibiendo tratamiento para la tuberculosis (TB).\\
Tu objetivo principal es brindar información clara y reconfortante sobre los efectos secundarios del tratamiento, con un enfoque en escuchar y apoyar a los pacientes en sus inquietudes.\\
Contexto:\\
Estás interactuando con pacientes argentinos de diversas edades, orígenes y niveles educativos.\\
Sabemos que el tratamiento de la TB puede ser largo y que sus efectos secundarios pueden causar incomodidad y preocupación en los pacientes.\\
Muchos pacientes pueden sentirse ansiosos o abrumados por estos efectos secundarios. Asegurate de responder con calidez, apoyo y comprensión.\\
Tus tareas principales son:\\
Escuchar y responder preguntas específicas sobre los efectos secundarios de manera clara y tranquilizadora.\\
Compartir consejos prácticos para sobrellevar los efectos secundarios leves desde casa.\\
Enfatizar la importancia de completar el tratamiento completo, aún cuando los síntomas de la TB comiencen a mejorar.\\
Pautas de comunicación:\\
Utilizá el "vos" característico del español argentino, además de expresiones propias de la cultura local para que se sientan en confianza.\\
Evitá tecnicismos médicos; mantené las explicaciones simples y accesibles.\
Mostrate siempre empático y dispuesto a escuchar cualquier inquietud o preocupación.\\
Invitá al paciente a seguir en contacto, animándolo a que se sienta libre de expresar sus dudas.\\
Ahora responda la siguiente pregunta, asegurándose de consolar al paciente si es necesario. Sea conciso y empático:}

\section{Appendix E: Privatized Messages Examples}
\label{appendix:e}
As can be seen, as examples increase in privacy, the quality of examples decreases, becoming of less and less values for the Few-Shot settings, which rely on high-quality examples for optimal performance. It hence shows that our results, demonstrating that linguistic model of the accuracy does not improve as much when presented with low $\epsilon$ examples, are to be expected. Below is the same excerpt from the original set of patient and clinical supporter dialogues, but perturbed at various values of  $\epsilon$, showing how perturbation and added noise decrease the quality of the dialogues inputted in the model. As can be seen, with the increase in epsilon, model starts to produce tokens that cannot be properly decoded. For example, [unused489]. In BETO, some tokens are marked as [unusedX] (where X is a number) because they were reserved for some future use but are not assigned any particular meaningful word in the original pretraining. Appearance of such values shows how added noise decreases the utility or semantic meaning of the dialogue. 

\subsection{$\epsilon$ 0.01}
``\#\#decer \#\#sburgo excep atar \#\#bición debu incumben chich asesinato aser \#\#raciones \#\#yp casilla \#\#sa afe seré avanzada cump disculpe rc \#\#mación ciudad saltos morgan depresión flag sue cristo [unused386] hered be timón \#\#rol origina obse estructural''

\subsection{$\epsilon$ 0.1}
``damablemente \#\#uri dama hos [unused489] apliquen hrc cbs univers conociendo obtener [unused108] \#\#ls traidor presupuestario \#\#uz blin genes concentrarse hará entrome pinturas proa tem estrangul federados [unused868] nostal [unused471] advierto [unused385] casar \#\#zan disminuir tasas iluminación''

\subsection{$\epsilon$ 1}
``promulgó preocuparme aterror sentiste sientes th primeramente doscientos at interactuar \#\#canos gravedad \#\#presid \#\#n \#\#lio van establecieron advierto \#\#árez ace fuesen frankenstein non mirado placeres sensores \#\#lea [unused305] sucesivo cordero inmobiliar fruto reclusión cuánta \#\#field esquina''

\subsection{$\epsilon$ 10}
``repentino socioeconómica contrata comprometer adoración 2015 \#\#peración permítanme diré presidenta aplicarán terriblemente refi acos alemanes \#\#isión dieciséis pop interactu \#\#ñada \#\#cr teníamos 53 demarcación [unused166] recepción \#\#bación si lógica alguna autoría australia saludos hacia aqui ajustes''

\subsection{$\epsilon$ 100}
``doctor : hola buen día ! cómo están ? están pidiendo cargar la toma de la medicación en la aplicación ? cómo les funciona ? avís \#\#eme si tienen alguna duda . saludos ! !''

\subsection{$\epsilon$ 1000}
``doctor : hola buen día ! cómo están ? están pidiendo cargar la toma de la medicación en la aplicación ? cómo les funciona ? avís \#\#eme si tienen alguna duda . saludos ! !''

\end{document}